\documentclass{article}

\usepackage{techreport}
\usepackage{geometry}
\usepackage{multirow}
\usepackage{graphicx}
\usepackage{multicol}
\usepackage{hyperref}
\usepackage{adjustbox}
\usepackage{caption}
\usepackage{subcaption}

\usepackage[utf8]{inputenc} 
\usepackage[T1]{fontenc}    
\usepackage{hyperref}       
\usepackage{url}            
\usepackage{booktabs}       
\usepackage{amsfonts}       
\usepackage{nicefrac}       
\usepackage{microtype}      
\usepackage{lipsum}
\usepackage{fancyhdr}       
\usepackage{graphicx}       
\graphicspath{{media/}}     
\usepackage{amsmath}
\usepackage{multirow}
\usepackage{multirow,makecell}

\title{\raisebox{-0.3\height}{\includegraphics[height=2.0em]{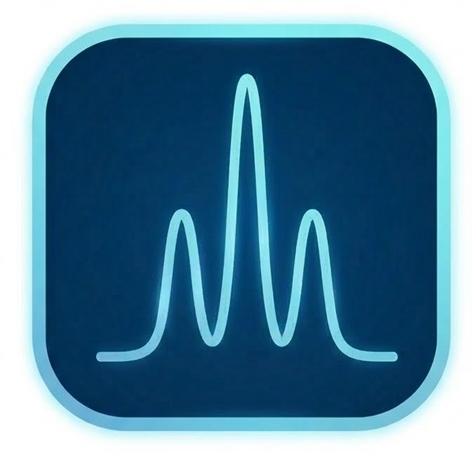}} ~SpecXMaster Technical Report}

\author{%
  DP Technology * \\\\
  March 25, 2026
}

\begin{document}
\maketitle

\vspace{-2.0em} 

\begin{center}
    \renewcommand{\arraystretch}{0.8}
    \begin{tabular}{c} 
        \href{https://specxmaster.bohrium.com}{
            \raisebox{-0.6ex}{\includegraphics[height=1.2em]{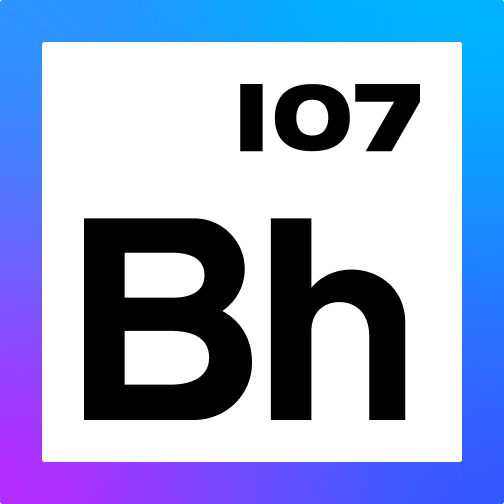}}
        \;\;\texttt{https:specxmaster.bohrium.com}
        } \\[0.2em]
    \end{tabular}
\end{center}

\begin{abstract}
Intelligent spectroscopy serves as a pivotal element in AI-driven closed-loop scientific discovery, functioning as the critical bridge between matter structure and artificial intelligence. However, conventional expert-dependent spectral interpretation encounters substantial hurdles, including susceptibility to human bias and error, dependence on limited specialized expertise, and variability across interpreters. To address these challenges, we propose SpecXMaster, an intelligent framework leveraging Agentic Reinforcement Learning (RL) for NMR molecular spectral interpretation. SpecXMaster enables automated extraction of multiplicity information from both ¹H and ¹³C spectra directly from raw FID (free induction decay) data. This end-to-end pipeline enables fully automated interpretation of NMR spectra into chemical structures. It demonstrates superior performance across multiple public NMR interpretation benchmarks and has been refined through iterative evaluations by professional chemical spectroscopists. We believe that SpecXMaster, as a novel methodological paradigm for spectral interpretation, will have a profound impact on the organic chemistry community.

\end{abstract}

\section{Introduction}
Nuclear Magnetic Resonance (NMR) spectroscopy stands as the cornerstone of structural chemistry, functioning as the primary analytical tool for the definitive identification and characterization of organic compounds. By providing an intricate map of molecular environments and connectivity, NMR spectra serve as the "molecular fingerprints" that enable researchers to validate synthetic outcomes and explore complex chemical spaces~\cite{Emwas_2020_10}. In the era of AI-driven closed-loop scientific discovery, the ability to rapidly and accurately interpret spectroscopic data has transitioned from a mere routine task to a critical bottleneck in the innovation pipeline~\cite{Tom_2024_08}. The integration of high-throughput experimental platforms with automated data acquisition systems has generated vast amounts of raw Free Induction Decay (FID) data, necessitating a paradigm shift in how these signals are translated into chemical knowledge. Consequently, the development of robust, intelligent methodologies for structural elucidation is no longemerely a technical pursut but a prerequisite for accelerating the pace of modern chemical research and bridging the gap between raw physical observationson and actionable molecular insights.

Despite the analytical power of NMR spectroscopy, translating spectral data into definitive molecular structures remains an inherently expert-dependent process. This reliance on human intervention introduces critical hurdles that impede the scalability of modern chemical research. Manual interpretation, as a deductive process, is highly susceptible to human bias and error, heavily influenced by a researcher's prior expectations and subjective experience. Even for seasoned spectroscopists, navigating congested spectral regions or interpreting complex splitting patterns is prone to cognitive fatigue and oversight, potentially leading to misassignments that can derail downstream synthetic efforts. Furthermore, the steep learning curve required to master NMR interpretation creates an "expertise gap," where the availability of high-level expertise becomes the scarcest resource in a laboratory. As high-throughput synthesis and automated sampling become standard, the human interpreter has emerged as the primary bottleneck; data acquisition can take minutes, whereas rigorous elucidation may require hours or even days~\cite{Granda_2018_07}. Compounding this issue is significant inter-interpreter variability, where the same set of ¹H and ¹³C spectra can yield divergent structural hypotheses depending on an expert's individual heuristic approach, thereby complicating data reproducibility and the standardization of chemical databases. In summary, the conventional manual workflow is increasingly incompatible with the demands of AI-driven, autonomous discovery cycles. To bridge this gap, transitioning from "expert-assisted" to "machine-autonomous" spectral interpretation is not merely an optimization but a structural necessity for the future of organic chemistry.

Computational methods have fundamentally revolutionized NMR spectroscopy, driving significant advancements in structural biology and modern chemistry~\cite{Das_2025_09}. Quantum chemical (QM) methods provide precise predictions of NMR parameters for detailed characterization, while machine learning (ML) techniques complement these by automating spectral assignments and predicting chemical shifts. The rapid advent of "SpectraML" (Spectroscopy Machine Learning) has catalyzed a shift from traditional expert-dependent workflows toward automated analysis capable of handling high-dimensional data~\cite{Guo_2025}. In forward tasks (molecule-to-spectrum prediction), frameworks such as NMRNet utilize SE(3) Transformers to model atomic environments~\cite{Xu_2024}. Furthermore, semi-supervised approaches have been proposed that leverage millions of unlabeled spectra for training, addressing the scarcity of peak-assigned NMR datasets~\cite{jin2026human}. The field has increasingly focused on inverse tasks (spectrum-to-molecule inference), where multitask learning frameworks combining CNNs and Transformers can predict molecular connectivity directly from 1D spectra, significantly reducing the structural search space~\cite{Hu_2024}. More recently, the emergence of multimodal Large Language Models (LLMs) such as SpectraLLM~\cite{Su_2025}, MolSpectLLM~\cite{Shen_2025}, and Spectro~\cite{Chacko_2024_11} has enabled joint reasoning across multiple spectroscopic modalities, including NMR, IR, and MS, to mimic the holistic approach of human experts. Furthermore, frameworks like NMR-Solver~\cite{jin2025nmr} integrate spectral-guided fragment optimization to enable interpretable reasoning and efficient refinement of molecular candidates, while DeepSPInN formulates elucidation as a Markov decision process (MDP) solved via deep reinforcement learning~\cite{Devata_2024_01}. While large-scale experimental databases like NMRexp now provide millions of records to train models~\cite{Wang_2025_12}, significant barriers to true automated intelligence persist.

Despite the rapid progress in computational spectroscopy, several critical gaps continue to hinder the practical deployment of AI in chemical research. Most existing frameworks suffer from an incomplete processing pipeline, frequently relying on simplified chemical shift lists or peak tables rather than directly interfacing with raw FID data, which remains a critical bottleneck for training truly robust and generalizable models. Furthermore, current methodologies often exhibit poor generalization when confronted with complex or novel molecules, frequently underperforming in real-world applications due to inherent algorithmic limitations and a heavy reliance on pre-existing structural databases or known structural information. Perhaps most significantly, these systems often lack the "intellectual" feedback and iterative reasoning capabilities of professional spectroscopists; they typically function as one-way mapping tools that neglect prior chemical knowledge and fail to perform the self-reflective logical deduction required to resolve discrepancies in experimental data.

To address these identified challenges, we propose SpecXMaster, an intelligent framework that introduces a novel methodological paradigm by leveraging Agentic RL for NMR spectral interpretation. Unlike traditional models that frequently rely on simplified chemical shift lists or peak tables, SpecXMaster employs an end-to-end pipeline that directly interfaces with raw FID data. By incorporating advanced signal processing techniques, the framework automates the extraction of quantitative peak parameters and multiplicity information from both NMR spectra of $^1$H and $^{13}$C.

\begin{figure}[ht]
\centering
\includegraphics[width=0.8\textwidth]{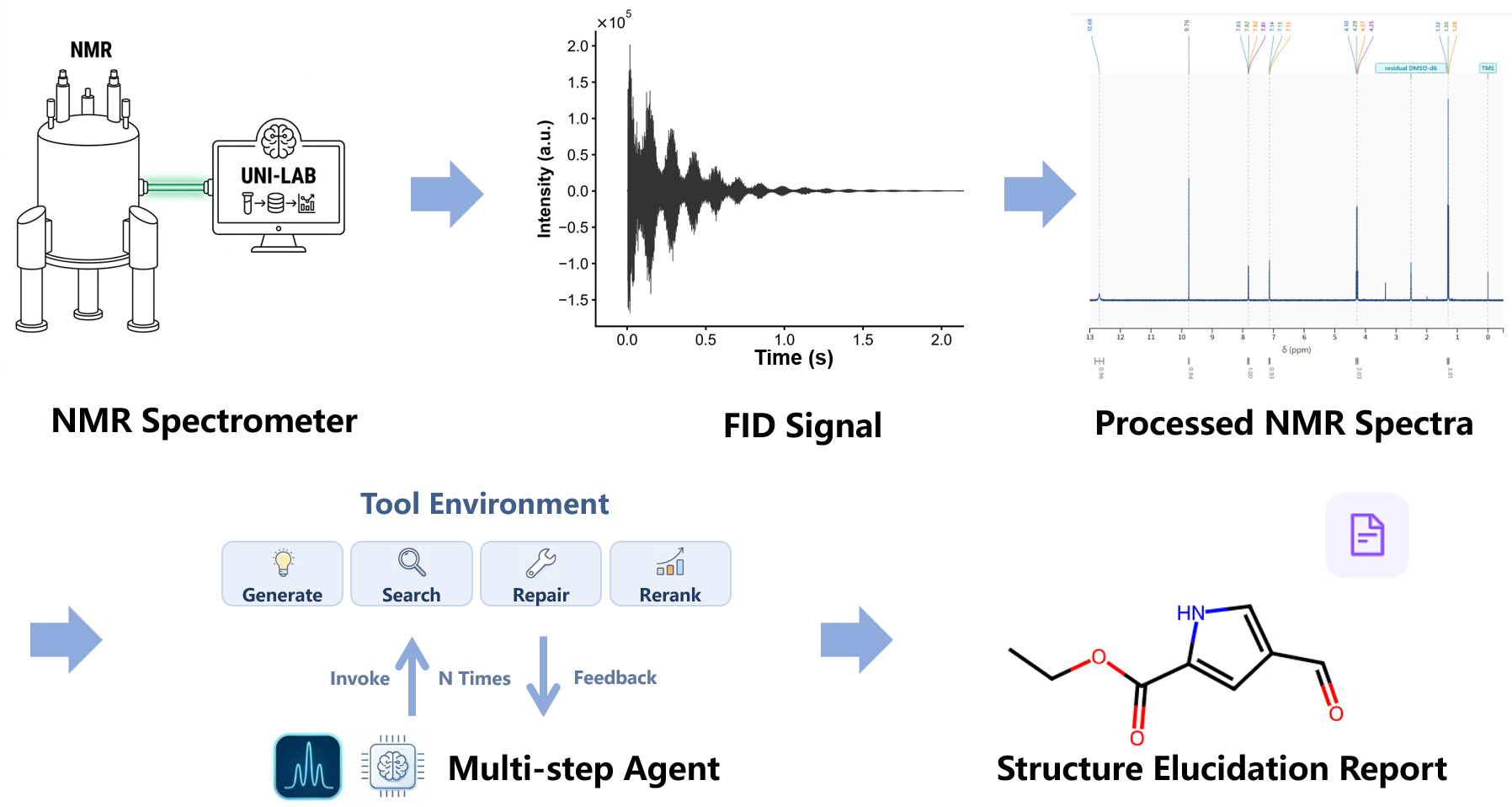}
\caption{End-to-end pipeline of SpecXMaster}\label{intro}
\end{figure}

As shown in \textbf{Figure 1} , SpecXMaster effectively translates complex physical signals into standardized, structured representations of NMR signals, which serve as the primary input for a Multi-step Agent that operates within a specialized tool environment featuring \textbf{Generate}, \textbf{Search}, \textbf{Repair}, and \textbf{Rerank} capabilities. By mimicking the iterative reasoning, discrepancy analysis, and error-correction logic of professional spectroscopists, SpecXMaster achieves an automated transition from raw NMR signals to precise chemical structures and comprehensive elucidation reports. This agentic approach moves beyond simple numerical error matching to assess overall structural plausibility, offering a robust, interpretable, and scalable solution for the organic chemistry community.

\section{Full Processing Steps from FID to NMR Spectrum}
\subsection{Overview}

NMR instruments record experimental signals in the form of \textbf{Free Induction Decay (FID)} time-domain data. Converting FID signals into interpretable spectra and structured chemical information is a critical step in automated NMR analysis.

We developed a \textbf{Python-based automated FID processing and spectral interpretation module} that transforms raw instrument output into:

\begin{itemize}
\item frequency-domain NMR spectra
\item peak lists and multiplet annotations
\item quantitative peak parameters
\item structured textual representations of NMR signals
\end{itemize}

This module serves as the \textbf{front-end spectral processing component} of the multi-agent NMR analysis system, bridging raw experimental data and downstream structure interpretation modules.

The pipeline integrates conventional digital signal processing methods with machine learning-based multiplet identification to enable robust spectral analysis across diverse spectra.

\subsection{System Architecture}

The automated FID processing system consists of four major stages:
\begin{enumerate}
\item FID preprocessing
\item Fourier transformation and spectrum generation
\item Peak detection and multiplet identification
\item Spectral annotation and text generation
\end{enumerate}

\begin{figure}[ht]
\centering
\includegraphics[width=0.8\textwidth]{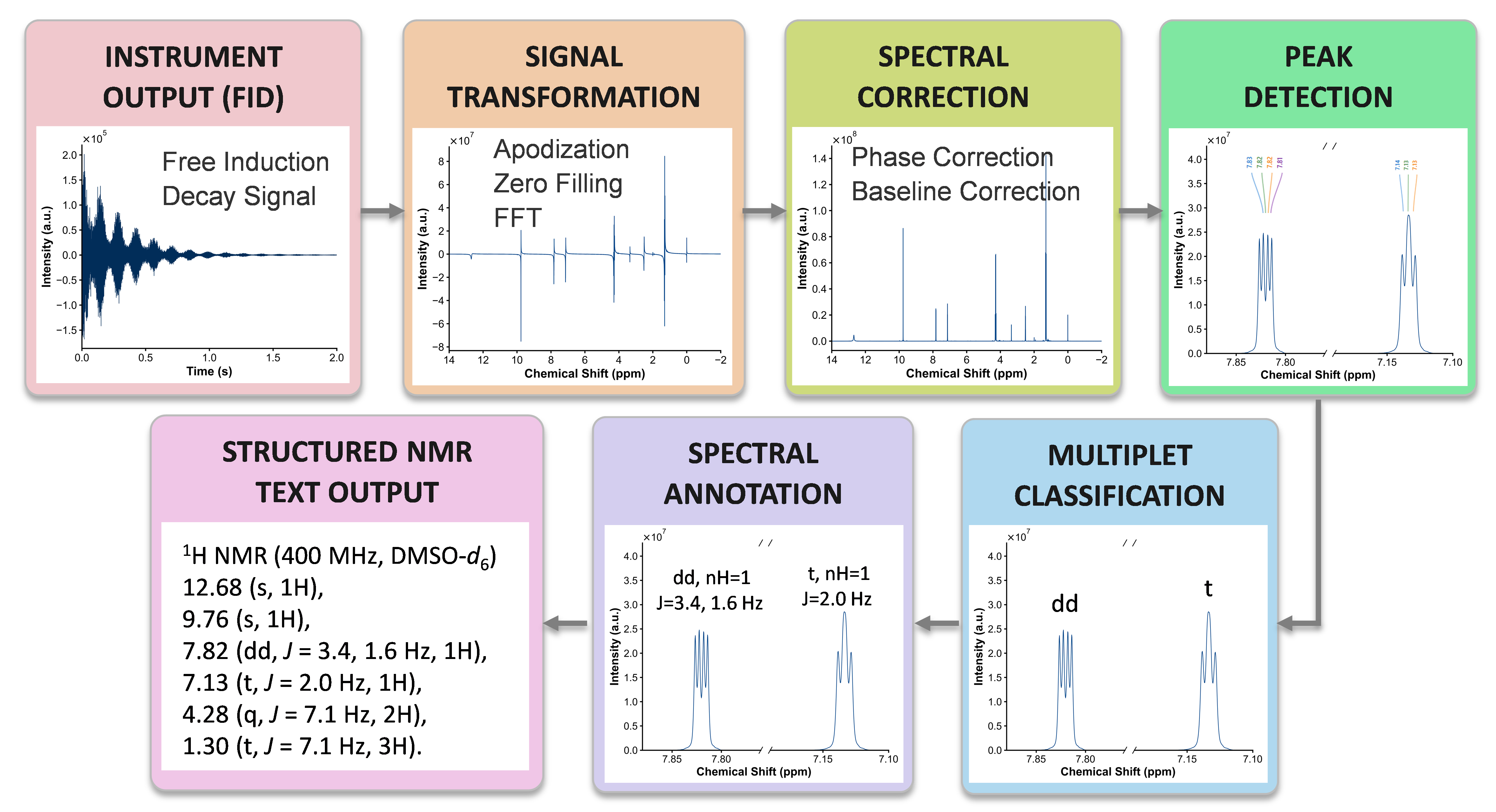}
\caption{Overview of Nuclear Magnetic Resonance (NMR) Data Processing}\label{fid_workflow}
\end{figure}

The system is implemented in Python using numerical computing libraries for spectral processing and neural network frameworks for multiplet classification.

\subsection{Signal Transformation}

The raw FID signal must first be transformed from the time domain into the frequency domain. This stage includes apodization, zero filling, and Fourier transformation. The raw FID data is read and processed using the nmrglue library~\cite{helmus2013nmrglue}.

\textbf{Apodization}: Apodization is applied to reduce truncation artifacts and improve the signal-to-noise ratio. The system supports several window functions commonly used in NMR signal processing~\cite{ernst1987principles}, including exponential multiplication, Gaussian weighting and sine-bell apodization. These window functions can be parameterized to balance spectral resolution and sensitivity depending on experimental conditions.

\textbf{Zero Filling}: Zero filling is used to increase digital resolution in the frequency domain~\cite{bartholdi1973fourier}. The original FID signal is extended by appending zeros until the signal length reaches a power-of-two value, which also improves the efficiency of the Fourier transform.

\textbf{Fourier Transformation}: After preprocessing, the time-domain FID signal $s(t)$ is converted into a frequency-domain spectrum using the Fast Fourier Transform (FFT)~\cite{cooley1965algorithm}:

\begin{equation}
S(\omega) = \mathcal{F}\{s(t)\}
\end{equation}

where $S(\omega)$ represents the frequency-domain spectral intensity. The resulting spectrum represents resonance signals as peaks located at characteristic chemical shift positions.

\subsection{Spectral Correction}

Following Fourier transformation, the spectrum typically contains distortions introduced during acquisition and signal processing. These artifacts must be corrected before reliable peak detection can be performed.

\textbf{Phase Correction}:Phase distortions arise from imperfections in receiver timing and instrument electronics. If uncorrected, these distortions cause asymmetric or dispersive peak shapes. Automatic phase correction is implemented based on the ACME algorithm~\cite{chen2002acme}, which optimizes zero-order and first-order phase parameters based on entropy minimization. The optimization objective aims to minimize spectral asymmetry and maximize peak sharpness.

\textbf{Baseline Correction}: Baseline distortions may arise from instrument drift, solvent suppression artifacts, or truncation effects. Baseline correction is performed using the polynomial fitting and the improved asymmetric least squares (IAsLS) algorithm combined with iterative masking of detected peaks using the pybaselines pack~\cite{pybaselines}. This procedure prevents the fitting process from being influenced by true resonance signals. After baseline correction, the spectrum exhibits a stable baseline suitable for quantitative peak analysis.

\textbf{Peak Detection}: Peak detection identifies candidate resonance signals from the processed spectrum~\cite{koradi1998autopsy}. The detection procedure consists of three main stages, including "noise level estimation", "local maxima detection" and "peak boundary determination". Noise levels are estimated from baseline regions of the spectrum. Peaks are identified based on signal-to-noise thresholds combined with local maximum detection. For each detected peak, the following parameters are extracted: chemical shift ($\delta$, ppm),  peak intensity, peak width, and integrated peak area. Overlapping signals are segmented using derivative-based boundary detection.

\subsection{Multiplet Identification and Spectral Annotation}

After peak detection, clusters of neighboring peaks are analyzed to determine multiplet patterns arising from spin--spin coupling. Rule-based multiplet identification often fails when signals overlap or when spectra contain distortions, while machine learning can effectively solve this problem~\cite{cobas2020nmr,jonas2019rapid}. To improve robustness, we implemented a \textbf{neural network-based multiplet classification model}. The model takes a localized spectral segment centered around candidate peaks as input and predicts the multiplet category.The supported multiplet types include:
singlet(s), doublet(d), triplet(t), quartet(q), multiplet(dd, td, dt, ddd......) and complex or overlapping patterns(m). The neural network was trained on a dataset of annotated spectra containing diverse multiplet patterns, which improves multiplet recognition accuracy, particularly in spectra containing partially overlapping signals.

Once peaks and multiplets are identified, the system generates annotated spectral information, including chemical shift values, multiplicity, integration values, and peak grouping information. Each detected signal is represented as a structured entry containing all extracted parameters.

\subsection{Structured NMR Text Generation}

To support downstream automated interpretation, spectral annotations are converted into standardized textual descriptions. A selected example output is illustrated:

\begin{minipage}{0.4\textwidth}
\begin{verbatim}
1H NMR (400 MHz, CDCl3): 7.26 (d, J = 8.4 Hz, 2H), 6.85 (d, J = 8.4 Hz, 2H), 
3.78 (s, 3H)
\end{verbatim}
\end{minipage}

This representation allows seamless integration with structure elucidation algorithms, chemical databases, and downstream reasoning modules.

\section{Agent Architecture}

\subsection{Overview}

\begin{figure}[ht]
    \centering
    \includegraphics[width=\linewidth]{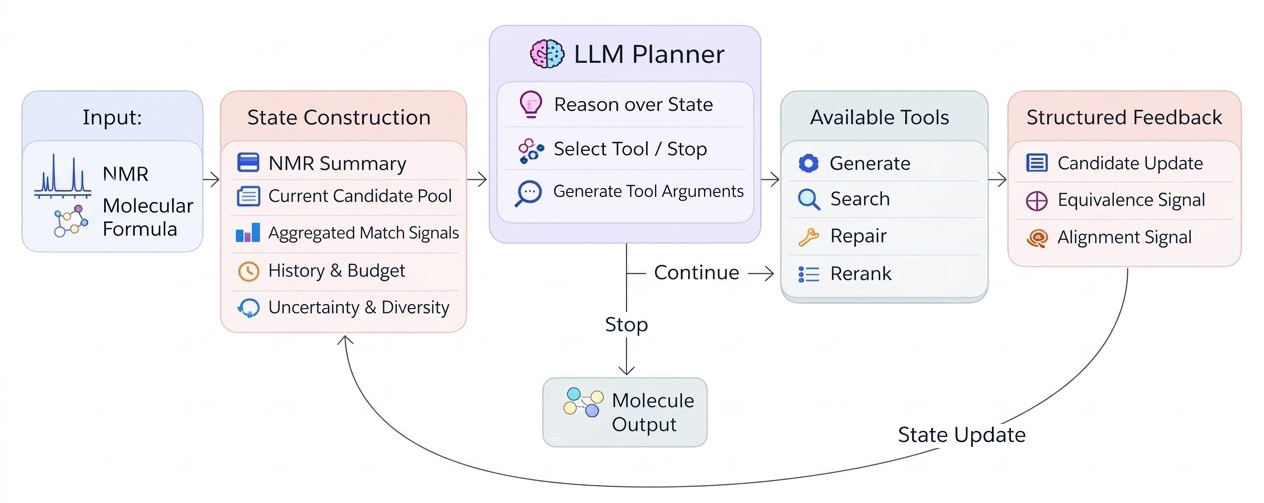}
    \caption{Overview of SpecXMaster, the proposed agentic framework for NMR-based molecular structure elucidation. The agent iteratively constructs a decision state, selects actions over the tool environment, receives structured feedback, and updates the state until termination.}
    \label{fig:workflow}
\end{figure}

Figure~\ref{fig:workflow} illustrates the overall workflow of SpecXMaster, our agentic framework for NMR-based structure elucidation~\cite{elyashberg2021case,nugroho2019computationally}. Starting from the input NMR observation, the system first constructs a compact decision state that summarizes the current reasoning context, including the NMR summary, the current candidate pool, aggregated matching signals, and trajectory-level information such as history, budget, uncertainty, and diversity. Based on this state, the agent policy selects the next operation and, when necessary, produces the corresponding tool arguments, following the general paradigm of tool-augmented language models that learn to invoke external functions in a context-dependent manner~\cite{schick2023toolformer,karpas2022mrkl}.

The available actions include candidate generation, candidate search, local repair, and reranking. After each action is executed, the environment returns structured feedback in the form of candidate updates, equivalence signals, and alignment signals. These feedback signals are then incorporated into the next-round state update, forming a closed-loop decision process. The interaction continues until the agent determines that the current candidate set is sufficiently good and outputs the final molecular structure.

Unlike one-shot structure prediction, SpecXMaster formulates NMR interpretation as an iterative reasoning-and-action process over a tool environment, where the model repeatedly evaluates the current state and determines the next step under intermediate feedback~\cite{karpas2022mrkl,yao2023react,shen2023hugginggpt}. The key role of the agent is therefore not to generate the final molecule in a single step, but to adaptively decide how to explore, refine, and evaluate candidate structures under intermediate feedback.

\subsection{Problem Formulation}

Formally, given an observation $x$, the agent interacts with a tool environment $\mathcal{T}$ for at most $T$ rounds. At round $t$, the policy produces a state-conditioned action
\begin{equation}
a_t \sim \pi_\theta(\cdot \mid s_t),
\label{eq:policy_main}
\end{equation}
where $s_t$ denotes the current reasoning state and $\pi_\theta$ denotes the agent policy.

Each action is factorized as
\begin{equation}
a_t = \big(a_t^{\mathrm{type}}, a_t^{\mathrm{arg}}\big),
\label{eq:action_factorization}
\end{equation}
where $a_t^{\mathrm{type}}$ specifies the molecular operation to invoke and $a_t^{\mathrm{arg}}$ specifies its corresponding arguments. In the current system,
\begin{equation}
a_t^{\mathrm{type}} \in \{\texttt{Generate},\ \texttt{Search},\ \texttt{Optimize},\ \texttt{ReRank},\ \texttt{Stop}\}.
\end{equation}

After executing $a_t$, the tool environment returns a structured observation
\begin{equation}
o_t = \mathcal{T}(a_t),
\end{equation}
and the next reasoning state is updated as
\begin{equation}
s_{t+1} = g(s_t, a_t, o_t),
\end{equation}
where $g$ is the state update operator.

The full reasoning trajectory is therefore
\begin{equation}
\tau = (x, s_1, a_1, o_1, s_2, a_2, o_2, \ldots, s_T, a_T, o_T).
\label{eq:trajectory}
\end{equation}

At termination, the system outputs the final molecular hypothesis
\begin{equation}
\hat{y} = \phi(C_T),
\label{eq:final_prediction}
\end{equation}
where $C_T$ denotes the terminal candidate pool and $\phi(\cdot)$ denotes the final selection rule.

\subsection{Tool Environment}
\label{sec:tool_environmet}

We define a unified tool environment $\mathcal{T}$ that exposes a small set of molecular operations to the agent. In the current framework, these operations include candidate generation, candidate search, local optimization, and final re-ranking. Rather than treating these tools as isolated modules, we regard them as the external action interface through which the agent explores and refines molecular hypotheses.

\paragraph{Generate.}
The \textit{Generate} tool employs an autoregressive model to generate end-to-end molecular structure from spectral inputs, inspired by the approach of NMRPeak~\cite{xu2026synergistic}. Concretely, the model predicts molecular SMILES sequences conditioned on the given NMR spectra. To obtain a diverse yet high-quality hypothesis set, beam search decoding is adopted to produce a ranked collection of candidate molecules. 

This generation stage provides the agent with an initial pool of chemically valid structures that are broadly consistent with the observed spectral patterns. The resulting candidate set serves as the starting point for subsequent tool interactions, including candidate search, local refinement, and final re-ranking, thereby enabling efficient exploration of the molecular hypothesis space.

\paragraph{Database search.}
The \textit{Search} tool enables rapid retrieval of spectrally similar molecules from a large-scale repository of known compounds. It leverages the database introduced in NMR-Solver~\cite{jin2025nmr}, which contains 106 million chemically valid molecules curated from the PubChem~\cite{kim2025pubchem} dataset. Each entry is associated with simulated $^1\mathrm{H}$ and $^{13}\mathrm{C}$ NMR spectra, providing a comprehensive reference space for efficient spectrum-based matching. Given an input NMR spectrum, the tool returns a ranked list of candidate molecular structures whose simulated spectra are most consistent with the query.

To achieve both efficiency and retrieval quality, the search is performed in two stages: an initial retrieval using vector representations with an HNSW~\cite{malkov2018efficient} index identifies candidate molecules, followed by re-ranking based on peak-level set similarity of the spectra. This hybrid strategy enables sub-second querying at the scale of hundreds of millions of compounds while maintaining high matching fidelity.

\paragraph{Repair.}
The \textit{Repair} tool is implemented on top of REINVENT4~\cite{loeffler2024reinvent4} and is used for targeted optimization of candidate molecules during the iterative reasoning process. Given an input candidate molecule $m$, the tool performs reinforcement-learning-based molecular optimization to improve its consistency with the observed spectrum, rather than generating a new structure from scratch, following the general paradigm of goal-directed molecular optimization from a starting compound~\cite{zhou2019moldqn,gow2022rlchemistry}.

Concretely, the repair objective is defined through a forward spectral prediction module developed in-house. Given a molecule $m$, the predictor produces its simulated spectrum $\hat{s}(m)$, which is then compared against the observed spectrum $s^{*}$ through a similarity function $\mathrm{Sim}(\hat{s}(m), s^{*})$. This similarity score is used as the reward signal:
\begin{equation}
R_{\mathrm{repair}}(m)=\mathrm{Sim}\!\left(\hat{s}(m), s^{*}\right).
\end{equation}

Starting from the current candidate, REINVENT4 optimizes the molecular policy toward candidates with higher spectral reward. The resulting tool is therefore particularly useful in cases where the current candidate pool already contains structurally plausible molecules, but additional local optimization is needed to improve spectrum-level agreement, consistent with prior work on constrained or starting-point-guided molecular optimization~\cite{he2024evaluation,zhou2019moldqn}.

\paragraph{Peak assignment.}

This \textit{Assign} tool addresses the problem of NMR peak assignment. Given a molecular structure and an experimental spectrum, the objective is to establish a correspondence between experimental signals and structurally defined sites. This is formulated as a structured relation
$ A \subseteq \mathcal{I} \times \mathcal{J}, $
where $\mathcal{I}$ denotes the set of structural sites and $\mathcal{J}$ denotes the set of observed spectral signals. The feasible assignment space is constrained by structural and observational consistency requirements $ A \in \mathcal{A}$.

In practice, the set of structural sites $\mathcal{I}$ is constructed through an equivalence analysis that groups atoms indistinguishable under symmetry or dynamic averaging. Since NMR experiments probe ensemble-averaged signals, assignment is performed at the level of equivalence groups rather than individual atoms.

The assignment is constructed based on multiple complementary sources of compatibility. For each potential pair $(i,j)$ between a structural site and an experimental signal, a compatibility function $\phi(i,j)$ is defined to characterize their agreement:
\begin{equation}
\phi(i,j) =
\phi_{\delta}(i,j)
\cdot
\phi_{\mathrm{disc}}(i,j)
\cdot
\phi_{J}(i,j).
\end{equation}
Here, $\phi_{\delta}$ measures the agreement between predicted and observed chemical shifts based on structure-derived shift estimates, 
$\phi_{\mathrm{disc}}$ encodes compatibility between discrete structural motifs and observed multiplicity patterns, and 
$\phi_{J}$ captures relational consistency arising from predicted scalar coupling relationships.
These components are constructed from empirical rules and physically motivated relationships and are interpreted as compatibility measures rather than exact predictions.

Chemical shifts are predicted using a structure-based deep learning model, trained on over a million reference spectra, providing expected reference values for each site~\cite{jin2026human}.
Coupling values are estimated by Karplus-type relations~\cite{karplus1959contact} of the form
\begin{equation}
J(\theta) = A \cos^2 \theta + B \cos \theta + C,
\end{equation}
where $\theta$ denotes the dihedral angle along the coupling pathway. These relations provide a physically grounded connection between molecular structure and observable coupling patterns, incorporating both empirical rules and geometric priors.

In addition, the assignment process explicitly accounts for non-ideal experimental conditions. The observed spectrum can be viewed as a partially observed realization of the underlying structural response:
\begin{equation}
X = g(c, A) + \varepsilon,
\end{equation}
where $g$ denotes the structural mapping to observable signals and $\varepsilon$ captures effects such as missing, merged, or weakly resolved signals. As a result, compatibility is evaluated in a tolerant manner, allowing deviations consistent with realistic measurement conditions.

Overall, peak assignment is formulated as a structured correspondence problem that integrates empirical knowledge, physical relations, and observational constraints to produce a globally consistent mapping between structural sites and experimental signals.

\paragraph{Candidate rerank.}

The \textit{Rerank} tool extends the peak assignment framework from a single structure to a candidate set. For each candidate structure $c \in \mathcal{C}$, a peak assignment procedure is first performed to establish a mapping between structural sites and experimental signals. Rather than comparing structures directly, candidates are evaluated based on the quality of their best achievable spectral explanation.

Formally, the score of a candidate structure is defined as the minimum assignment cost over all feasible assignment configurations:
\begin{equation}
S(c) = \min_{A \in \mathcal{A}(c)} \mathcal{L}(A \mid c, X),
\end{equation}
where $X$ denotes the experimental spectrum, $A$ represents a valid assignment configuration, and $\mathcal{A}(c)$ is the set of all feasible assignments under structural and observational constraints.

The total cost function is composed of multiple components capturing different aspects of assignment quality:
\begin{equation}
\mathcal{L}(A \mid c, X)
=
\lambda_{\mathrm{match}} \mathcal{L}_{\mathrm{match}}
+
\lambda_{\mathrm{cov}} \mathcal{L}_{\mathrm{coverage}}
+
\lambda_{\mathrm{obs}} \mathcal{L}_{\mathrm{observation}}.
\end{equation}

The matching term $\mathcal{L}_{\mathrm{match}}$ measures local agreement between assigned pairs of structural sites and experimental signals:
\[
\mathcal{L}_{\mathrm{match}}
=
\sum_{(i,j)\in A} \ell(i,j),
\]
where $\ell(i,j)$ encodes deviations in continuous attributes and inconsistencies in discrete structural patterns. The coverage term penalizes incomplete explanations of the spectrum:
\begin{equation}
\mathcal{L}_{\mathrm{coverage}}
=
\alpha \, N_{\mathrm{unassigned\;sites}}
+
\beta \, N_{\mathrm{unexplained\;signals}}.
\end{equation}
In addition, an observation term $\mathcal{L}_{\mathrm{observation}}$ accounts for non-ideal experimental effects through controlled tolerance of missing, merged, or weakly resolved signals.

Importantly, the assignment $A$ is obtained under global consistency constraints, so that the optimization is not a sum of independent local matches but a structured matching problem over the entire spectrum. The resulting score therefore reflects the best self-consistent explanation of the observed data by a given candidate.

The final prediction is obtained by selecting the candidate with the lowest assignment cost:
\begin{equation}
c^\star = \arg\min_{c \in \mathcal{C}} S(c).
\end{equation}

\paragraph{Hard-case processor.}

As part of the \textit{ReRank} module, we introduce a hard-case processor based on hyperbolic representation learning to improve discrimination on challenging near-tie candidates.
This design is inspired by prior work on hyperbolic representation learning, which has shown that hyperbolic geometry is particularly effective for modeling hierarchical or fine-grained relational structure in representation space~\cite{nickel2017poincare,nickel2018lorentz,ganea2018hyperbolic}. It is also motivated by recent findings that hyperbolic scoring can improve difficult ranking problems with small margins between highly similar candidates~\cite{wu2026beyond}.
The key idea is that hyperbolic geometry provides larger effective separation in crowded local neighborhoods, making it more suitable for distinguishing small-margin candidates than conventional Euclidean similarity.

Let $\mathbf{z}^{(s)} \in \mathbb{R}^{d}$ denote the projected spectrum representation and $\mathbf{z}^{(m)}_i \in \mathbb{R}^{d}$ denote the projected representation of the $i$-th candidate molecule after integrating molecular and prior information. We use Uni-Mol~\cite{zhou2023uni} for molecular representations and BART~\cite{lewis2020bart} for spectral representations, followed by NMRPeak. We map these Euclidean vectors into the Lorentz model of hyperbolic space,
\begin{equation}
\mathbb{H}^{d}_{\kappa}
=
\left\{
\mathbf{x}\in\mathbb{R}^{d+1}
\;\middle|\;
\langle \mathbf{x},\mathbf{x}\rangle_{\mathcal{L}}=-\kappa,\ x_0>0
\right\},
\end{equation}
where $\kappa>0$ is the curvature radius and $\langle \cdot,\cdot\rangle_{\mathcal{L}}$ is the Lorentz inner product
\begin{equation}
\langle \mathbf{x},\mathbf{y}\rangle_{\mathcal{L}}
=
-x_0 y_0 + \sum_{j=1}^{d} x_j y_j.
\end{equation}

We first obtain normalized Euclidean projections
\begin{equation}
\tilde{\mathbf{z}}^{(s)}=\mathrm{LN}(\mathbf{z}^{(s)}), \qquad
\tilde{\mathbf{z}}^{(m)}_i=\mathrm{LN}(\mathbf{z}^{(m)}_i),
\end{equation}
and then lift them to hyperbolic points using the exponential map at the origin:
\begin{equation}
\mathbf{h}^{(s)}=\exp_{\mathbf{o}}^{\kappa}(\tilde{\mathbf{z}}^{(s)}), \qquad
\mathbf{h}^{(m)}_i=\exp_{\mathbf{o}}^{\kappa}(\tilde{\mathbf{z}}^{(m)}_i),
\end{equation}
where $\mathbf{o}=(\sqrt{\kappa},0,\dots,0)$ is the hyperbolic origin. The hyperbolic geodesic distance between the spectrum and the $i$-th candidate is then
\begin{equation}
d_{\mathbb{H}}\!\left(\mathbf{h}^{(s)},\mathbf{h}^{(m)}_i\right)
=
\sqrt{\kappa}\,
\operatorname{arcosh}\!\left(
-\frac{\langle \mathbf{h}^{(s)},\mathbf{h}^{(m)}_i\rangle_{\mathcal{L}}}{\kappa}
\right).
\end{equation}

We define the hard-case score using the negative geodesic distance
\begin{equation}
s_i^{\mathrm{hard}}
=
-\,d_{\mathbb{H}}\!\left(\mathbf{h}^{(s)},\mathbf{h}^{(m)}_i\right),
\end{equation}
so that candidates more consistent with the spectrum receive larger scores.

To further emphasize hard-case discrimination, we optimize the processor with a hybrid objective that combines hyperbolic contrastive learning and margin-based hard-negative ranking:
\begin{equation}
\mathcal{L}_{\mathrm{hard}}
=
\mathcal{L}_{\mathrm{con}}
+
\lambda_{\mathrm{rank}} \mathcal{L}_{\mathrm{rank}}
+
\lambda_{\mathrm{reg}} \mathcal{L}_{\mathrm{reg}},
\end{equation}
where $\lambda_{\mathrm{rank}}$ and $\lambda_{\mathrm{reg}}$ balance the contributions of the auxiliary objectives.

The contrastive term is defined as
\begin{equation}
\mathcal{L}_{\mathrm{con}}
=
-\log
\frac{
\exp\!\left(s_{i^{+}}^{\mathrm{hard}}/\eta\right)
}{
\exp\!\left(s_{i^{+}}^{\mathrm{hard}}/\eta\right)
+
\sum_{i^{-}\in\mathcal{N}_{\mathrm{hard}}}
\exp\!\left(s_{i^{-}}^{\mathrm{hard}}/\eta\right)
},
\end{equation}
where $\eta>0$ is a temperature parameter and $\mathcal{N}_{\mathrm{hard}}$ denotes the mined hard-negative set.

To further enlarge the separation margin between the positive candidate and hard negatives, we introduce a ranking term
\begin{equation}
\mathcal{L}_{\mathrm{rank}}
=
\sum_{i^{-}\in\mathcal{N}_{\mathrm{hard}}}
\left[
m - s_{i^{+}}^{\mathrm{hard}} + s_{i^{-}}^{\mathrm{hard}}
\right]_{+},
\end{equation}
where $m>0$ is a ranking margin and $[\cdot]_{+}=\max(\cdot,0)$.

Finally, to stabilize the learned representation space, we use a regularization term
\begin{equation}
\mathcal{L}_{\mathrm{reg}}
=
\left\|\tilde{\mathbf{z}}^{(s)}\right\|_2^2
+
\left\|\tilde{\mathbf{z}}^{(m)}_{i^{+}}\right\|_2^2
+
\sum_{i^{-}\in\mathcal{N}_{\mathrm{hard}}}
\left\|\tilde{\mathbf{z}}^{(m)}_{i^{-}}\right\|_2^2 .
\end{equation}

This objective simultaneously improves global separability, enforces a larger margin on hard negatives, and stabilizes the local geometry of the learned embedding space. The resulting hard-case score is then used as a dedicated signal for difficult candidates before downstream ranking.

\subsection{State Representation and Structured Feedback}

The agent operates on a compact state representation that summarizes the current progress of the reasoning trajectory. Rather than reprocessing raw spectral signals at every round, the policy reasons over a state $s_t$ that captures both the current hypothesis set and the interaction context. Specifically, at round $t$, the state contains the current candidate pool $C_t$, candidate-level summary signals, equivalence-related judgments over the current hypotheses, as well as the action history and the remaining interaction budget.

After executing an action $a_t$, the tool environment returns a structured observation rather than raw tool outputs. We use a compact feedback interface so that the policy can reason over high-level progress signals instead of low-level execution details. In the current framework, the observation is summarized as
\begin{equation}
o_t = \big(c_t,\ e_t,\ m_t\big),
\label{eq:feedback_tuple}
\end{equation}
where $c_t$ denotes the \textbf{candidate summary}, $e_t$ denotes the \textbf{equivalence judgment}, and $m_t$ denotes the \textbf{alignment information}. Here, $c_t$ summarizes the updated status of the candidate pool, $e_t$ indicates whether multiple candidates correspond to equivalent or near-equivalent structural hypotheses, and $m_t$ reflects how well the current hypothesis set agrees with the spectral and chemistry-aware evaluation signals.



\section{Agentic Reinforcement Learning Framework}
\label{sec:agentic_rl_framework}

\subsection{Policy Formulation}

Based on the interaction process defined in Eqs.~\eqref{eq:policy_main}--\eqref{eq:final_prediction}, we model the agent as a state-conditioned stochastic policy over the tool environment~\cite{sutton2018reinforcement}. Using the action factorization in Eq.~\eqref{eq:action_factorization}, the policy is written as
\begin{equation}
\pi_\theta(a_t \mid s_t)
=
\pi_\theta\!\left(a_t^{\mathrm{type}}, a_t^{\mathrm{arg}} \mid s_t\right),
\label{eq:policy_factorized}
\end{equation}
where $s_t$ denotes the current reasoning state and $a_t$ denotes the next molecular action.

The policy backbone is implemented by an LLM, which serves as a structured controller that selects molecular operations and instantiates their arguments, rather than acting as a generic free-form text generator~\cite{schick2023toolformer,li2023apibank,patil2024gorilla}. At each round, it is responsible for selecting which molecular operation to invoke next, producing the corresponding arguments when needed, and determining whether the current reasoning process should terminate.

The policy is optimized by maximizing the expected return of the full reasoning process:
\begin{equation}
\max_\theta \ \mathbb{E}_{\tau \sim \pi_\theta}\!\left[\sum_{t=1}^{T} r_t\right],
\label{eq:rl_objective_step}
\end{equation}
where $r_t$ is the step-level reward assigned after executing action $a_t$ and observing feedback $o_t$.

This formulation is suitable for NMR-based structure elucidation because the usefulness of an action is often delayed and depends on subsequent refinement steps.




\subsection{Reward Design}

Given the reasoning trajectory in Eq.~\eqref{eq:trajectory}, we assign a step-level reward after each action--observation pair $(a_t, o_t)$. The reward at step $t$ is defined as
\begin{equation}
r_t
=
\lambda_{\mathrm{fmt}} r_t^{\mathrm{fmt}}
+
\lambda_{\mathrm{eff}} r_t^{\mathrm{eff}}
+
\lambda_{\mathrm{tool}} r_t^{\mathrm{tool}}
+
\lambda_{\mathrm{align}} r_t^{\mathrm{align}},
\label{eq:step_reward}
\end{equation}
where each term captures one aspect of desirable agent behavior.

\paragraph{Format reward.}
Let $\mathcal{V}_{\mathrm{fmt}}$ denote the set of valid and executable action outputs. We define
\begin{equation}
r_t^{\mathrm{fmt}}
=
\mathbf{1}\!\left[a_t \in \mathcal{V}_{\mathrm{fmt}}\right].
\end{equation}

\paragraph{Efficiency reward.}
Let $T_0$ denote a target reasoning budget. We define
\begin{equation}
r_t^{\mathrm{eff}}
=
-\,\mathbf{1}[t > T_0].
\end{equation}

\paragraph{Tool-usage reward.}
Let $\mathcal{V}_{\mathrm{tool}}$ denote the set of successful tool invocations, $\mathcal{P}_{\mathrm{tool}}$ denote the set of productive tool outcomes, and $\mathcal{F}_{\mathrm{tool}}$ denote the set of failed or invalid tool calls. The tool-usage reward is defined as
\begin{equation}
r_t^{\mathrm{tool}}
=
\alpha_{\mathrm{succ}} \, \mathbf{1}\!\left[(a_t,o_t)\in\mathcal{V}_{\mathrm{tool}}\right]
+
\alpha_{\mathrm{prog}} \, \mathbf{1}\!\left[(a_t,o_t)\in\mathcal{P}_{\mathrm{tool}}\right]
-
\alpha_{\mathrm{fail}} \, \mathbf{1}\!\left[(a_t,o_t)\in\mathcal{F}_{\mathrm{tool}}\right],
\end{equation}
where $\alpha_{\mathrm{succ}}, \alpha_{\mathrm{prog}}, \alpha_{\mathrm{fail}} > 0$ are tunable coefficients.

\paragraph{Alignment reward.}
Given the structured feedback in Eq.~\eqref{eq:feedback_tuple}, let $r_{\mathrm{cand}}(o_t)$ measure improvement in candidate summary and let $r_{\mathrm{eq}}(o_t)$ measure reduction of redundant or unstable hypotheses. We define
\begin{equation}
r_t^{\mathrm{align}}
=
\beta_1 \, r_{\mathrm{cand}}(o_t)
+
\beta_2 \, r_{\mathrm{eq}}(o_t),
\end{equation}
where $\beta_1,\beta_2 \ge 0$ are weighting coefficients.

Accordingly, the total return of a reasoning process is defined as the sum of step-level rewards along the trajectory:
\begin{equation}
R(\tau)=\sum_{t=1}^{T} r_t.
\label{eq:return_from_step_rewards}
\end{equation}

\subsection{Training Strategy}

We adopt a two-stage training strategy. In the first stage, the policy is initialized with supervised fine-tuning (SFT) or behavior-cloned warm-start data so that the model can learn the required action format and interact with the tool environment in a stable and executable manner. In the second stage, we further optimize the initialized policy using Group Relative Policy Optimization (GRPO)~\cite{shao2024deepseekmath, zhao2025molreasoner, li2025agentplanning}.

For each training instance, we sample a group of responses
\begin{equation}
\{y_i\}_{i=1}^{G}, \qquad y_i \sim \pi_{\theta_{\mathrm{old}}}(\cdot \mid x),
\end{equation}
where each response $y_i$ corresponds to one complete rollout under the current policy and therefore induces a full reasoning process $\tau_i$. For each sampled response, we execute the corresponding multi-step interaction in the tool environment and compute its total return
\begin{equation}
R_i = R(\tau_i)=\sum_{t=1}^{T_i} r_{i,t}.
\end{equation}

Following the group-relative formulation, we normalize rewards within each sampled group to obtain the relative advantage:
\begin{equation}
\hat{A}_i
=
\frac{R_i - \mu_R}{\sigma_R + \epsilon},
\qquad
\mu_R = \frac{1}{G}\sum_{j=1}^{G}R_j,
\qquad
\sigma_R = \sqrt{\frac{1}{G}\sum_{j=1}^{G}(R_j-\mu_R)^2}.
\end{equation}
Here, $\epsilon$ is a small constant for numerical stability.

Let $y_{i,t}$ denote the $t$-th optimized token in response $y_i$. Define the policy ratio
\begin{equation}
\rho_{i,t}(\theta)
=
\frac{\pi_\theta(y_{i,t}\mid x, y_{i,<t})}
{\pi_{\theta_{\mathrm{old}}}(y_{i,t}\mid x, y_{i,<t})}.
\end{equation}
Then the GRPO objective is
\begin{equation}
\mathcal{L}_{\mathrm{GRPO}}(\theta)
=
- \frac{1}{G}\sum_{i=1}^{G}\frac{1}{|y_i|}
\sum_{t=1}^{|y_i|}
\min\!\Big(
\rho_{i,t}(\theta)\hat{A}_i,\,
\mathrm{clip}\big(\rho_{i,t}(\theta), 1-\epsilon_c, 1+\epsilon_c\big)\hat{A}_i
\Big),
\label{eq:grpo_objective}
\end{equation}
where $\epsilon_c$ is the clipping coefficient.

\section{Experiment}
\subsection{Experimental Setup}

\paragraph{Baselines and compared methods.}
We compare SpecXMaster against three categories of methods. The first category is the standalone generation tool, which corresponds to the \textit{Generate} module in our tool environment (Section~\ref{sec:tool_environmet}). In this setting, molecular structures are predicted directly from spectral inputs without iterative multi-step decision making, and the resulting candidate list is used as a generation-based baseline.

The second category consists of workflow baselines. These baselines preserve the same overall tool environment, prompt format, and multi-step interaction protocol as SpecXMaster, but replace the trained agent policy with a general-purpose foundation model. In our experiments, we instantiate two workflow baselines using GPT-5.2~\cite{openai2025gpt52} and Qwen2.5-7B~\cite{yang2024qwen25}, respectively. This comparison is intended to isolate the value of the learned agent policy from the value of the tool environment itself.

The final category is the proposed SpecXMaster model. It is built on the same Qwen2.5-7B backbone as the corresponding workflow baseline, but uses a trained agent policy optimized through supervised fine-tuning and reinforcement learning (Section~\ref{sec:agentic_rl_framework}). Unless otherwise specified, all compared methods are evaluated under the same benchmark setting and input modes.




\paragraph{Benchmark setup, data split, and training data construction.}
We follow the benchmark construction and data split settings of NMRexp dataset~\cite{Wang_2025_12} described in Xu et al.~\cite{xu2026synergistic} to ensure fair and direct comparison with prior work. The overall statistics of the NMRexp dataset across different splits are summarized in Table~\ref{tab:nmrexp_stats}. The dataset is divided into training, validation, and test sets, each containing spectrum--structure pairs. 

\begin{table}[!htbp]
\centering
\caption{Statistics of the NMRexp dataset across different splits and spectrum types.}
\label{tab:nmrexp_stats}
\begin{tabular}{lcccc}
\toprule
Split & Total & $^{13}$C spectra & $^{1}$H spectra & Joint ($^{13}$C \& $^{1}$H) \\
\midrule
Train & 920,796 & 786,240 & 805,224 & 670,668 \\
Validation & 48,463 & 41,383 & 42,366 & 35,286 \\
Test & 107,696 & 91,830 & 94,163 & 78,297 \\
\midrule
Overall & 1,076,955 & 919,453 & 941,753 & 784,251 \\
\bottomrule
\end{tabular}
\end{table}



\paragraph{Training protocol.}
We adopt a two-stage training strategy, following the common practice of first performing supervised fine-tuning (SFT) to establish the desired output format and basic behavior, and then further improving the policy with reinforcement learning~\cite{ouyang2022instructgpt}. Since the generation tool has already been trained on the original training split, all agent-training data are constructed from the validation split of NMRexp, while the test split is reserved exclusively for evaluation. In the first stage, we perform SFT on 2{,}000 samples drawn from the validation split to teach the model the required structured output format and basic tool-use behavior. The SFT set consists of 1{,}000 Joint samples, 500 $^{13}$C spectra samples, and 500 $^{1}$H spectra samples. In the second stage, starting from the SFT-initialized checkpoint, we perform online reinforcement learning on the full validation split. This design avoids overlap with the generator's original training data and ensures that the agent is optimized on a split whose candidate structures are more consistent with the data distribution encountered during agent training.

\paragraph{Optimization details.}
For RL training, we use a batch size of 32 and train for 900 optimization steps. Unless otherwise specified, the same backbone model, tool environment, and evaluation protocol are used across all agent variants. 

\paragraph{Evaluation metrics.}
\label{par:evaluation_metrics}

Model performance is evaluated using the rank-matcher hit@$k$ metric. For each test sample $i$, the model produces a ranked list of candidate molecules
$\{\hat{y}_{i,1},\ldots,\hat{y}_{i,k}\}$.
A candidate is counted as correct when its rank-matcher key is identical to that of the ground-truth molecule.

The rank-matcher key is constructed by parsing the SMILES with RDKit, removing atom-map annotations and explicit hydrogen atoms, and generating a canonical isomeric representation. For molecules containing a single tetrahedral stereocenter, the configuration of that center is ignored. For molecules with multiple tetrahedral stereocenters, globally inverted enantiomers are treated as equivalent while relative stereochemistry is preserved. Unspecified stereogenic C=C bonds are normalized to the trans configuration, and E/Z annotations are ignored for terminal-nitrogen alkene patterns. Molecular connectivity must otherwise match exactly, and empty or chemically invalid predictions are counted as incorrect.

Let $\operatorname{key}_{\mathrm{rank}}(\cdot)$ denote the resulting molecular comparison key. The hit@$k$ metric is defined as

\begin{equation}
\mathrm{hit@}k
=
\frac{1}{N}
\sum_{i=1}^{N}
\mathbf{1}
\left[
\exists\, j \leq k:
\operatorname{key}_{\mathrm{rank}}(\hat{y}_{i,j})
=
\operatorname{key}_{\mathrm{rank}}(y_i)
\right],
\end{equation}

where $N$ is the number of test samples, $\hat{y}_{i,j}$ is the candidate ranked at position $j$ for sample $i$, $y_i$ is the corresponding ground-truth molecule, and $\mathbf{1}[\cdot]$ is the indicator function.

\subsection{Main Benchmark Results}





\begin{table*}[t]
\centering
\scriptsize
\setlength{\tabcolsep}{3.6pt}
\renewcommand{\arraystretch}{1.15}
\resizebox{\textwidth}{!}{
\begin{tabular}{lcccccccccccc}
\toprule
\multirow{2}{*}{Method}
& \multicolumn{3}{c}{hit@1}
& \multicolumn{3}{c}{hit@3}
& \multicolumn{3}{c}{hit@5}
& \multicolumn{3}{c}{hit@10} \\
\cmidrule(lr){2-4}
\cmidrule(lr){5-7}
\cmidrule(lr){8-10}
\cmidrule(lr){11-13}
& Joint & $^{13}$C spectra & $^{1}$H spectra
& Joint & $^{13}$C spectra & $^{1}$H spectra
& Joint & $^{13}$C spectra & $^{1}$H spectra
& Joint & $^{13}$C spectra & $^{1}$H spectra \\
\midrule

Generation Tool
& 0.707 & 0.479 & 0.490
& 0.813 & 0.592 & 0.616
& 0.840 & 0.630 & \textbf{0.658}
& 0.865 & \textbf{0.669} & \textbf{0.696} \\

Workflow (GPT-5.2)
& 0.651 & 0.433 & 0.437
& 0.746 & 0.529 & 0.545
& 0.766 & 0.555 & 0.573
& 0.780 & 0.568 & 0.591 \\

Workflow (Qwen2.5-7B)
& 0.625 & 0.416 & 0.419
& 0.717 & 0.507 & 0.522
& 0.736 & 0.532 & 0.549
& 0.749 & 0.544 & 0.566 \\

SpecXMaster
& \textbf{0.783} & \textbf{0.515} & \textbf{0.511}
& \textbf{0.895} & \textbf{0.620} & \textbf{0.630}
& \textbf{0.912} & \textbf{0.641} & 0.652
& \textbf{0.925} & 0.661 & 0.693 \\

\bottomrule
\end{tabular}
}
\caption{Main benchmark. We report hit@1, hit@3, hit@5, and hit@10 under three input modes: Joint ($^{13}$C \& $^{1}$H), $^{13}$C spectra, and $^{1}$H spectra. The best result in each column is highlighted in bold.}
\label{tab:main_benchmark}
\end{table*}

We first compare SpecXMaster with the standalone Generation Tool and two workflow baselines on the main benchmark. Following the rank-matcher evaluation protocol defined in \hyperref[par:evaluation_metrics]{Evaluation metrics}, we report hit@1, hit@3, and hit@5 under three spectrum settings: Joint, $^{13}$C spectra, and $^{1}$H spectra. The results are summarized in Table~\ref{tab:main_benchmark}.

The most important observation is that SpecXMaster consistently achieves the strongest top-ranked performance among all compared systems. On hit@1, SpecXMaster reaches 0.783 under the Joint setting, 0.515 with $^{13}$C spectra, and 0.511 with $^{1}$H spectra. Compared with the strongest workflow baseline, Workflow (GPT-5.2), these results represent absolute improvements of 0.132, 0.082, and 0.074, respectively. SpecXMaster also achieves the best hit@3 performance across all three settings, reaching 0.895, 0.620, and 0.630, respectively. These results indicate that the advantage of SpecXMaster does not arise simply from access to the same tool environment, but from a policy explicitly trained for NMR-guided multi-step decision making. The learned agent is better able to determine when to generate, search, repair, and rerank candidate structures based on intermediate feedback.

A second notable observation is that both workflow baselines consistently underperform the standalone Generation Tool across all spectrum settings and ranking cutoffs. For example, under the Joint setting, the Generation Tool achieves hit@1, hit@3, and hit@5 scores of 0.707, 0.813, and 0.840, whereas Workflow (GPT-5.2) reaches only 0.651, 0.746, and 0.766. Workflow (Qwen2.5-7B) performs further below the Generation Tool, with corresponding scores of 0.625, 0.717, and 0.736. Similar trends are observed for the two single-nucleus settings. This suggests that directly integrating a general-purpose language model into the tool environment is insufficient for this task. Although GPT-5.2 and Qwen2.5-7B are capable general-purpose reasoners, they do not natively possess the domain-aligned inductive bias required for NMR interpretation or directly model the statistical regularities of spectrum-to-structure mapping. Consequently, a prompting-based decision layer may introduce additional noise rather than reliably improving the candidate ranking.

Compared with the standalone Generation Tool, SpecXMaster provides substantial improvements in early precision. At hit@1, SpecXMaster improves from 0.707 to 0.783 under the Joint setting, from 0.479 to 0.515 with $^{13}$C spectra, and from 0.490 to 0.511 with $^{1}$H spectra. The improvement remains consistent at hit@3, where SpecXMaster raises performance from 0.813 to 0.895, from 0.592 to 0.620, and from 0.616 to 0.630 across the three settings, respectively. At hit@5, SpecXMaster improves from 0.840 to 0.912 under the Joint setting and from 0.630 to 0.641 with $^{13}$C spectra. For $^{1}$H spectra, the Generation Tool retains a small advantage at hit@5, achieving 0.658 compared with 0.652 for SpecXMaster. Overall, SpecXMaster achieves the strongest hit@1 performance in every spectrum setting and the best result in eight of the nine reported columns, demonstrating the effectiveness of combining spectrum-conditioned generation with learned candidate refinement and chemistry-aware ranking.

\subsubsection{Effect of RL-based Agent Optimization}

We evaluate the contribution of reinforcement learning by comparing three agent variants under the same tool environment: a fixed workflow baseline implemented with GPT-5.2, an SFT-only agent, and the final RL-optimized agent. Final structure prediction is evaluated using the rank-matcher hit@1 criterion defined in \hyperref[par:evaluation_metrics]{Evaluation metrics}.

Table~\ref{tab:sft_vs_rl_main} shows that reinforcement learning consistently improves the agent beyond its SFT initialization. SpecXMaster (RL) outperforms SpecXMaster (SFT) in final structure prediction under all three spectrum settings. Specifically, hit@1 increases from 0.745 to 0.783 under the Joint setting, from 0.500 to 0.515 with $^{13}$C spectra, and from 0.474 to 0.511 with $^{1}$H spectra. The corresponding absolute improvements are 0.038, 0.015, and 0.037, respectively. Although the improvement is relatively modest for the $^{13}$C-only setting, the consistent gains across all three settings indicate that RL improves the quality of the final candidate selection beyond supervised fine-tuning alone.

RL produces particularly large improvements in Format Validity. The SFT policy achieves validity scores of 0.768, 0.673, and 0.700 under the Joint, $^{13}$C, and $^{1}$H settings, respectively. After RL optimization, these scores increase to 0.999, 1.000, and 0.999. This result indicates that RL not only improves final structure prediction but also substantially stabilizes the agent's interaction behavior, including tool invocation, response formatting, and termination decisions.

The improvement in Case judgement further demonstrates that RL changes the agent's sequential decision strategy rather than merely refining surface-level output quality. Compared with the SFT policy, the RL policy improves Case judgement from 0.316 to 0.890 under the Joint setting, from 0.710 to 0.854 with $^{13}$C spectra, and from 0.802 to 0.918 with $^{1}$H spectra. The largest improvement occurs under the Joint setting, where the absolute gain reaches 0.574. These results suggest that the RL-trained agent is substantially better at determining whether the current candidate pool is sufficient or whether additional generation, search, repair, or reranking steps are required.

Finally, the fixed GPT-5.2 workflow remains below both learned agents across all reported metrics and spectrum settings. For example, its hit@1 scores are 0.651, 0.433, and 0.437, compared with 0.783, 0.515, and 0.511 for SpecXMaster (RL). Similar gaps are observed in Format Validity and Case judgement. This comparison indicates that the gains do not arise merely from access to the same tools. Instead, they result from learning a domain-aligned policy for using those tools adaptively, with RL providing further improvements in prediction accuracy, behavioral stability, and sequential decision making.

\begin{table*}[t]
\centering
\scriptsize
\setlength{\tabcolsep}{3.6pt}
\renewcommand{\arraystretch}{1.12}
\resizebox{\textwidth}{!}{
\begin{tabular}{lccccccccc}
\toprule
\multirow{2}{*}{Method}
& \multicolumn{3}{c}{hit@1}
& \multicolumn{3}{c}{Format Validity}
& \multicolumn{3}{c}{Case judgement} \\
\cmidrule(lr){2-4}
\cmidrule(lr){5-7}
\cmidrule(lr){8-10}
& Joint & $^{13}$C spectra & $^{1}$H spectra
& Joint & $^{13}$C spectra & $^{1}$H spectra
& Joint & $^{13}$C spectra & $^{1}$H spectra \\
\midrule
Workflow (GPT-5.2)
& 0.651 & 0.433 & 0.437
& 0.744 & 0.648 & 0.676
& 0.301 & 0.689 & 0.781 \\

SpecXMaster (SFT)
& 0.745 & 0.500 & 0.474
& 0.768 & 0.673 & 0.700
& 0.316 & 0.710 & 0.802 \\

SpecXMaster (RL)
& \textbf{0.783} & \textbf{0.515} & \textbf{0.511}
& \textbf{0.999} & \textbf{1.000} & \textbf{0.999}
& \textbf{0.890} & \textbf{0.854} & \textbf{0.918} \\
\bottomrule
\end{tabular}
}
\caption{
Comparison between the fixed workflow baseline, the SFT-only agent, and the RL-optimized agent. Final structure accuracy is measured using rank-matcher hit@1. The best result in each column is highlighted in bold.
}
\label{tab:sft_vs_rl_main}
\end{table*}

\subsubsection{Reward Ablation}

\begin{table*}[t]
\centering
\scriptsize
\setlength{\tabcolsep}{3.6pt}
\renewcommand{\arraystretch}{1.12}
\resizebox{\textwidth}{!}{
\begin{tabular}{lccccccccc}
\toprule
\multirow{2}{*}{Variant}
& \multicolumn{3}{c}{hit@1}
& \multicolumn{3}{c}{Avg. \# steps}
& \multicolumn{3}{c}{Useful action rate} \\
\cmidrule(lr){2-4}
\cmidrule(lr){5-7}
\cmidrule(lr){8-10}
& Joint & $^{13}$C spectra & $^{1}$H spectra
& Joint & $^{13}$C spectra & $^{1}$H spectra
& Joint & $^{13}$C spectra & $^{1}$H spectra \\
\midrule
Full reward
& \textbf{0.783} & \textbf{0.515} & \textbf{0.511}
& 4.18 & 4.96 & 4.71
& 0.352 & 0.331 & 0.339 \\

w/o efficiency reward
& 0.780 & 0.504 & 0.496 
& 5.34 & 6.52 & 6.08
& 0.309 & 0.286 & 0.293 \\

w/o tool-usage reward
& 0.777 & 0.497 & 0.495
& 4.33 & 5.05 & 4.80
& 0.288 & 0.272 & 0.279 \\

w/o alignment reward
& 0.771 & 0.494 & 0.489 
& \textbf{4.15} & \textbf{4.88} & \textbf{4.62}
& \textbf{0.356} & \textbf{0.335} & \textbf{0.342} \\
\bottomrule
\end{tabular}
}
\caption{Reward ablation results. We report hit@1, average number of reasoning steps, and useful action rate under three spectrum settings.}

\label{tab:reward_ablation_main}
\end{table*}

We next analyze the contribution of the step-level reward design by removing one reward component at a time, while keeping the extraction module and the default format reward enabled in all settings. The results in Table~\ref{tab:reward_ablation_main} show that the full reward delivers the best overall trade-off between final prediction quality, reasoning efficiency, and action quality.

\paragraph{Efficiency reward.}
When the efficiency reward is omitted, the average number of reasoning steps increases substantially across all three spectrum settings, confirming that this term is the primary mechanism for controlling trajectory length. Meanwhile, the accompanying decline in useful action rate indicates that many of the extra steps are unproductive, reflecting redundant exploration rather than effective progress.

\paragraph{Tool-usage reward.}
Without the tool-usage reward, the useful action rate drops consistently, together with a slight degradation in hit@1. This suggests that the tool-usage reward helps the policy learn when invoking a tool is genuinely beneficial, rather than merely encouraging more frequent interactions with the tool environment.

\paragraph{Alignment reward.}
Ablating the alignment reward yields the largest reduction in hit@1 across all three spectrum settings, even though the average number of reasoning steps becomes slightly smaller and the useful action rate remains competitive. This suggests that the alignment reward is the most critical component for improving final prediction quality, as it directly encourages the policy to steer the candidate pool toward better spectral consistency.

\subsection{Comprehensive Comparison on an Open 500-Case Evaluation Set}
\label{sec:open500_comparison}

To further assess the practical robustness of SpecXMaster under a lightweight and reproducible evaluation setting, we conduct an additional model-level comparison on FairNMR500, an open 500-case evaluation set sampled from the NMRexp test split in Table~\ref{tab:nmrexp_stats}. The dataset and evaluation code are publicly available on Hugging Face.\footnote{\url{https://huggingface.co/datasets/dptech/FairNMR500}}

For each example, the input contains the molecular formula, the $^{13}$C NMR peak list, and the $^{1}$H NMR peak list. For the direct-prompt baselines, we use a unified prompt template that asks the model to infer the molecular structure solely from the provided spectroscopic evidence and return a SMILES string in a fixed answer block. The full prompt template is provided in Appendix~\ref{app:open500_prompt}.

We report the top-1 hit rate over the 500 cases. For SpecXMaster, the top-ranked structure produced by the complete inference pipeline is used as the final prediction. All methods are evaluated using the rank-matcher hit@$k$ protocol defined in \hyperref[par:evaluation_metrics]{Evaluation metrics}, with $k=1$. Thus, a prediction is counted as correct when its rank-matcher molecular key is identical to that of the ground-truth structure; empty or chemically invalid predictions are counted as incorrect.

Formally, let $\operatorname{key}_{\mathrm{rank}}(\cdot)$ denote the normalized molecular key produced by this procedure. The hit rate is computed as
\begin{equation}
\mathrm{HitRate}_{\mathrm{rank}}
=
\frac{1}{500}
\sum_{i=1}^{500}
\mathbf{1}
\left[
\operatorname{key}_{\mathrm{rank}}(\hat{y}_i)
=
\operatorname{key}_{\mathrm{rank}}(y_i)
\right]
\times 100\% ,
\end{equation}
where $\hat{y}_i$ denotes the predicted SMILES for the $i$-th example, $y_i$ denotes the corresponding ground-truth SMILES, and $\mathbf{1}[\cdot]$ is the indicator function.

\begin{table}[!htbp]
\centering
\caption{
Comprehensive comparison on FairNMR500, an open 500-case evaluation set sampled from the NMRexp test split. Hit rate is computed using the rank-matcher criterion adopted by the reranking module, including its canonicalization and stereochemical-equivalence rules.
}
\label{tab:open500_comparison_rank_matcher}
\begin{tabular}{lcc}
\toprule
Method & Correct / Total & Hit rate (\%) \\
\midrule
SpecXMaster & 391 / 500 & \textbf{78.20} \\
GPT-6-astra + Codex & 381 / 500 & 76.20 \\
GPT-6-astra & 316 / 500 & 63.20 \\
Gemini 3.1 Pro Preview & 224 / 500 & 44.80 \\
DeepSeek V4 Pro & 191 / 500 & 38.20 \\
Claude Opus 4.7 & 175 / 500 & 35.00 \\
Qwen3.8-Max & 165 / 500 & 33.00 \\
GPT-5.5 & 145 / 500 & 29.00 \\
GPT-5.6-sol & 135 / 500 & 27.00 \\
Claude Opus 4.8 & 117 / 500 & 23.40 \\
\bottomrule
\end{tabular}
\end{table}

As shown in Table~\ref{tab:open500_comparison_rank_matcher}, SpecXMaster achieves the highest hit rate of 78.20\% (391/500), outperforming GPT-6-astra with Codex (76.20\%) and GPT-6-astra (63.20\%). Other frontier models, including Gemini 3.1 Pro Preview (44.80\%) and DeepSeek V4 Pro (38.20\%), achieve substantially lower performance. These results highlight the effectiveness of SpecXMaster for NMR-based molecular structure elucidation.


\section{Case Study}
To demonstrate the practical utility and robustness of the SpecXMaster framework, we conducted a comprehensive case study on \textbf{Ethyl 4-formyl-1H-pyrrole-2-carboxylate}, a representative heterocyclic compound in synthetic organic chemistry. This case study illustrates the seamless transition from raw physical signals to a refined chemical structure.(\textbf{Figure 4})
\begin{figure}[h]
\centering
\includegraphics[width=0.8\textwidth]{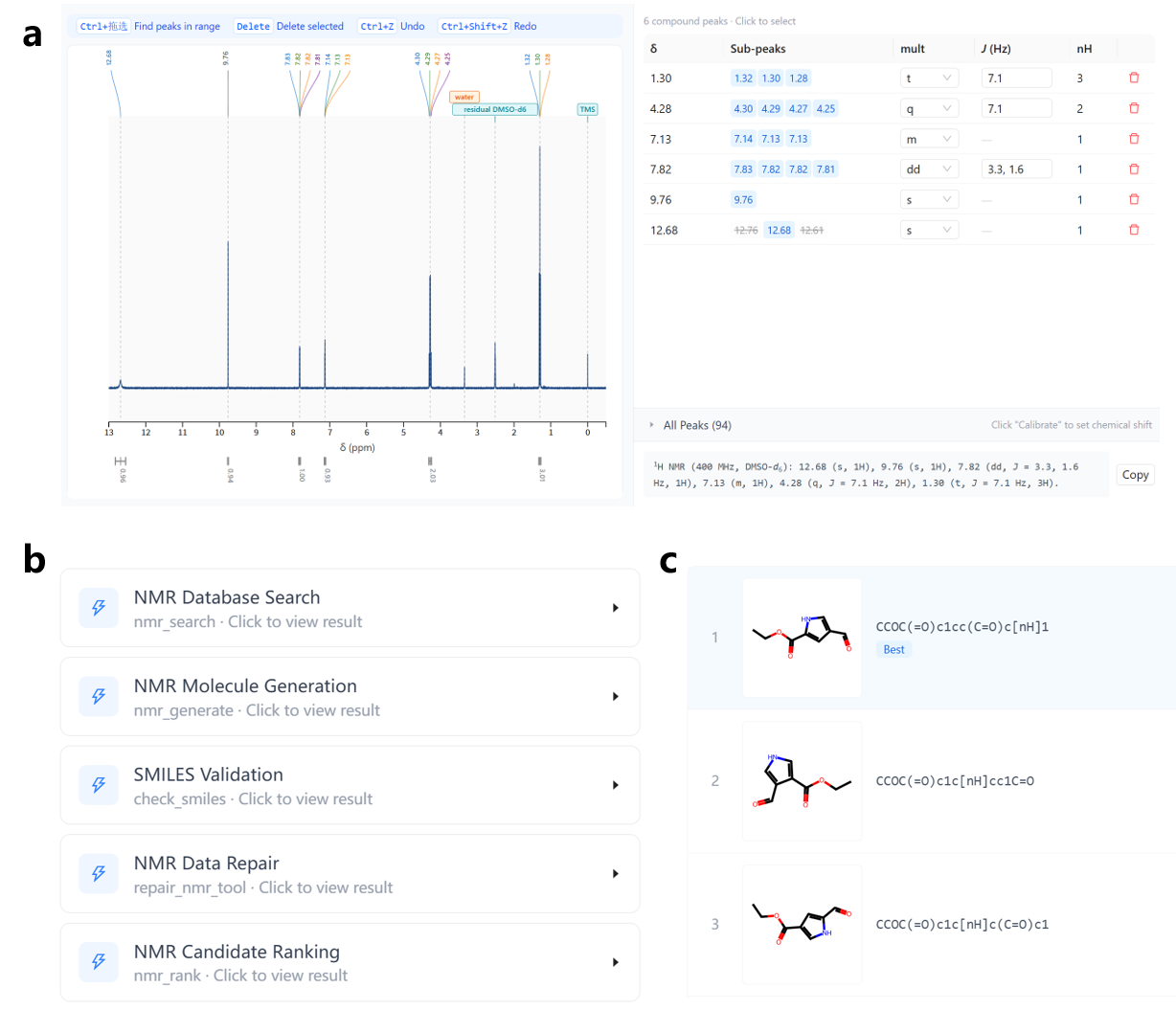}
\caption{A comprehensive case study of elucidating molecular structure from FID data: (a) transformation of FID data into spectrum and multiplicity analysis; (b) agentic reasoning for structural elucidation; (c) reranking of candidate structures and final structure identification.}\label{case_study}
\end{figure}
\subsection{Stage I: Signal Processing and Feature Extraction}
The process initiated with the acquisition of raw FID data from both $^1\text{H}$ and $^{13}\text{C}$ NMR experiments. SpecXMaster’s signal processing module directly performed automated Fourier transformation, phase correction, and baseline adjustment.The system successfully extracted quantitative multiplicity information, translating the complex spectral waves into a standardized textual format. For the $^1\text{H}$ and $^{13}\text{C}$ NMR spectrum, the extracted multiplicity text were as follows:

$^1\text{H}$ NMR(400 MHz, $DMSO{-}d_{6}$): $\delta$ 12.69 (s, 1H), 9.77 (s, 1H),  7.82 (dd, $J=3.4, 1.6$ Hz, 1H), 7.13 (dd, $J=2.1, 1.8$ Hz, 1H), 4.23 (q, $J=7.1$ Hz, 2H), 1.30 (t, $J=7.0$ Hz, 3H). $^{13}\text{C}$ NMR(101 MHz, $DMSO{-}d_{6}$): $\delta$ 185.8, 160.1, 131.1, 126.7, 124.6, 113.1, 60.3, 14.2. 

By converting the raw signals into this intermediate "Multiplicity Text," the framework effectively bridged the gap between physical data and chemical logic, ensuring that no critical coupling information was lost during preprocessing.

\subsection{Stage II: Agentic Reasoning and Structural Elucidation}
In the second stage, SpecXMaster performs structure elucidation through an agent-controlled tool-use process over the candidate space. Starting from the multiplicity text and molecular formula, the agent first uses \textbf{Generate} to propose an initial set of candidate structures. In this case, the correct molecule is not produced directly, but the generated candidates provide a reasonable starting point for subsequent exploration. The agent then invokes \textbf{Search} to expand the candidate pool with structurally related molecules from external databases. Although this step still does not return the exact target structure, it introduces candidates with a more plausible core scaffold, which provides a much stronger basis for downstream refinement. Next, the agent applies \textbf{Repair} to modify locally inconsistent structural fragments and correct scaffold-level details, gradually moving the candidate pool closer to the ground-truth molecule. Finally, \textbf{Rerank} is used to reorder the refined candidates according to spectrum--structure consistency, so that the correct structure can be prioritized at the top of the final list. This example illustrates the core advantage of the agentic framework: rather than relying on a single-step prediction, the agent progressively improves candidate quality through generate--search--repair--rerank interactions until the final structure is recovered. Other selected examples and their top-k reranked result are shown in \textbf{Appendix B}.

\section{Future Outlook}

\subsection{Towards Multi-dimensional and Multi-modal Structural Elucidation}
The future roadmap of SpecXMaster envisions a transition from an NMR-centric framework to a comprehensive multi-dimensional and multi-modal elucidation platform capable of addressing more complex chemical challenges. A primary focus will be the deep integration of 2D NMR spectroscopy, including HSQC, HMBC, COSY, and NOESY. By leveraging the high-order atomic connectivity and spatial proximity constraints provided by 2D spectra as inputs for the Agentic RL model, the system will significantly enhance its accuracy in identifying complex organic molecules, natural products, and subtle structural isomers. Furthermore, to meet the demands of real-world research, SpecXMaster will incorporate a specialized module for mixture analysis and quantitative yield determination. Through advanced spectral deconvolution techniques, the platform will not only achieve qualitative identification of multiple components but also perform quantitative NMR (qNMR) analysis to directly calculate product purity and reaction yields. By synergizing these capabilities with other modalities such as Infrared (IR) functional group features and Mass Spectrometry (LC-MS/GC-MS) fragmentation patterns, the system will emulate the holistic reasoning of human experts. This integration of reactant fragments and reaction context will allow the framework to perform cross-validation even in cases of signal overlap or missing data, establishing a closed-loop cognitive process from raw signals to precise structural determination.
\subsection{Platform Ecosystem and UniLab OS Integration}
The scalability and practical utility of SpecXMaster will be expanded through its integration into a broader, automated laboratory ecosystem.
\vspace{-1pt}
\textbf{Closed-Loop Autonomous Elucidation via UniLab OS}: To bridge the gap between digital decision-making and embodied experimentation, we will integrate SpecXMaster with UniLab OS, an AI-native operating system for autonomous laboratories~\cite{gao2025unilabos}. Using UniLab's  \textit{Action/Resource/Action\&Resource}(A/R/A\&R) model and transactional protocols, the platform will enable for a seamless "dry-wet" closed loop. This integration facilitates direct instrument connection for end-to-end structural analysis, allowing agents to autonomously trigger re-acquisition or experimental validation to resolve structural discrepancies.
\vspace{-1pt}
\textbf{Collaborative Ecosystem and Customized Development}: SpecXMaster aims to foster a shared intelligent elucidation ecosystem by collaborating with leading academic research groups and industrial partners. This collaboration will drive the co-creation of standardized datasets and the advancement of domain-specific models. Additionally, we will provide customized model development tailored to proprietary customer datasets, offering bespoke solutions that address unique chemical spaces and specific industrial requirements.
\vspace{-1pt}
\section{Conclusion}
In this work, we presented SpecXMaster, a novel intelligent framework that redefines the paradigm of NMR spectral interpretation through Agentic RL. By incorporating FID processing into an end-to-end pipeline, SpecXMaster overcomes the critical limitations of conventional expert-dependent methods, which are often hindered by human bias, variability, and the scarcity of specialized expertise. A key strength of our approach is the ability to interface directly with raw FID data rather than relying on simplified peak tables, ensuring that no critical spectral information is lost during the interpretation process. Through iterative reasoning and the integration of a multi-tool environment—including candidate generation, database search, and physics-guided local repair—SpecXMaster mimics the self-reflective logical deduction of professional spectroscopists. Experimental results demonstrate that SpecXMaster significantly outperforms existing one-shot models and general-purpose LLM, particularly in handling complex structural elucidation and "hard-case" candidates through hyperbolic representation learning. As a robust and scalable solution, SpecXMaster not only bridges the gap between raw physical signals and actionable molecular insights but also serves as a foundational component for the future of AI-driven, closed-loop scientific discovery in organic chemistry.





\bibliographystyle{unsrt}  
\bibliography{references}  

\newpage
\appendix

\section{Authorship and Acknowledgments}

Please cite this work as “DP Technology (2026)”. 

Correspondence regarding this technical report can be sent to jixh@dp.tech, gaozf@dp.tech \\

\begin{multicols}{2}
\raggedcolumns

\textbf{Core Contributor} \\
Yutang Ge \\
Yaning Cui \\
Hanzheng Li \\
Jun-Jie Wang \\
Fanjie Xu \\

\textbf{Contributors} \\
Jinhan Dong \\
Yongqi Jin \\
DongXu Cui \\
Peng Jin \\
Guojiang Zhao \\
Hengxing Cai \\
Tianci Yangfeng \\
Xueqing Chen \\
Hongshuai Wang \\
Changhao Nai \\
Jiarun Liu \\
Zheng Zhang

\columnbreak

\textbf{Project Lead} \\
Xiaohong Ji \\
Zhifeng Gao \\

\textbf{Team Management} \\
Rong Zhu \\
Linfeng Zhang \\

\end{multicols}

\section{Selected Examples of Internal Case Study}

Based on feedback from Professor Rong Zhu's group during SpecXMaster's internal testing, the table below presents the top-k ranking of candidate SMILES paired with their real structures, generated from inputs of multiplicity text and molecular formula

\begin{center}
\begin{tabular}{l c r}
    \toprule
    SMILES & Top-k \\
    \midrule
    C/C(C1=CC=CC=C1)=C\verb|\|CC(OC2=CC=C(C(OC)=O)C=C2)C3=CC=CC=C3&1 \\
    Cl[C@H]1CCCC2=CC=CC=C21 & 1 \\
    O=C(/C(C/C(C(C1=CC=CC=C1)=O)=C/NC2=CC=CC=C2)=C\verb|\|NC3=CC=CC=C3)C4=CC=CC=C4  & 1 \\
    OC(C=C1C)=CC(C)=C1C2=NC(C3=CC=CC(C4=C(C)C=C(O)C=C4C)=N3)=CC=C2  & 1 \\
    O=C(OC(C)(C)C)N1CCC(Cl)C1 & 1 \\
    O=C(C1=C(N)SC(C2=CC=C3C=CC=CC3=C2)C1)OCC & 1 \\
    O=C1N(CC(C)=O)C(C2=CC=CC=C21)=O & 1 \\
    CC(C)(O)C1=CC=C(C\verb|#|N)C=C1 & 1 \\
    C=C=C1CC(C(=O)OCc2ccccc2)=C(OCc2ccccc2)O1 & 1 \\
    O=c1[nH]c(-c2ccccc2)co1 & 1 \\
    COc1ccccc1CN1CCN(Cc2ccccc2OC)C1 & 1 \\
    CC(=CC1CN(Cc2ccccc2)CCN1Cc1ccccc1)c1ccccc1 & 1 \\
    C\verb|#|CC1(c2cccc(Cl)c2)COC(N)=C1C\verb|#|N& 1 \\
    O=C(C(C=CC=C1)=C1C2=O)N2OCC3=CSC=N3& 1 \\
    C\verb|#|CC1=CN(S(C2=CC=C(C)C=C2)(=O)=O)C3=C1C=CC(Br)=C3 & 1 \\
    CC(=O)Nc1cccc(C=O)c1 & 1 \\
    OC(C(C1C(C(OC)=O)=C(C(OC)=O)C2(O1)C3=CC=CC=C3)=C2C4)CN4S(C5=CC=C(C)C=C5)(=O)=O& 2 \\
    O=C(C1=CC=CS1)OCCC/C=C/CCCOC(C2=CC=CS2)=O & 3 \\
    CC1(C)CC=C(OS(=O)(C(F)(F)F)=O)C([Si](C)(C)C)=C1 & 4 \\
    F/C(N1C(C=C(C)C(C)=C2)=C2N=N1)=C\verb|\|C3=CC=C(C(OC)=O)C=C3 & 4 \\
    \bottomrule
\end{tabular}
\end{center}









\clearpage
\section{Detailed Structure Elucidation Report}
\label{app:detailed_report}

\begin{center}
\includegraphics[width=0.8\textwidth]{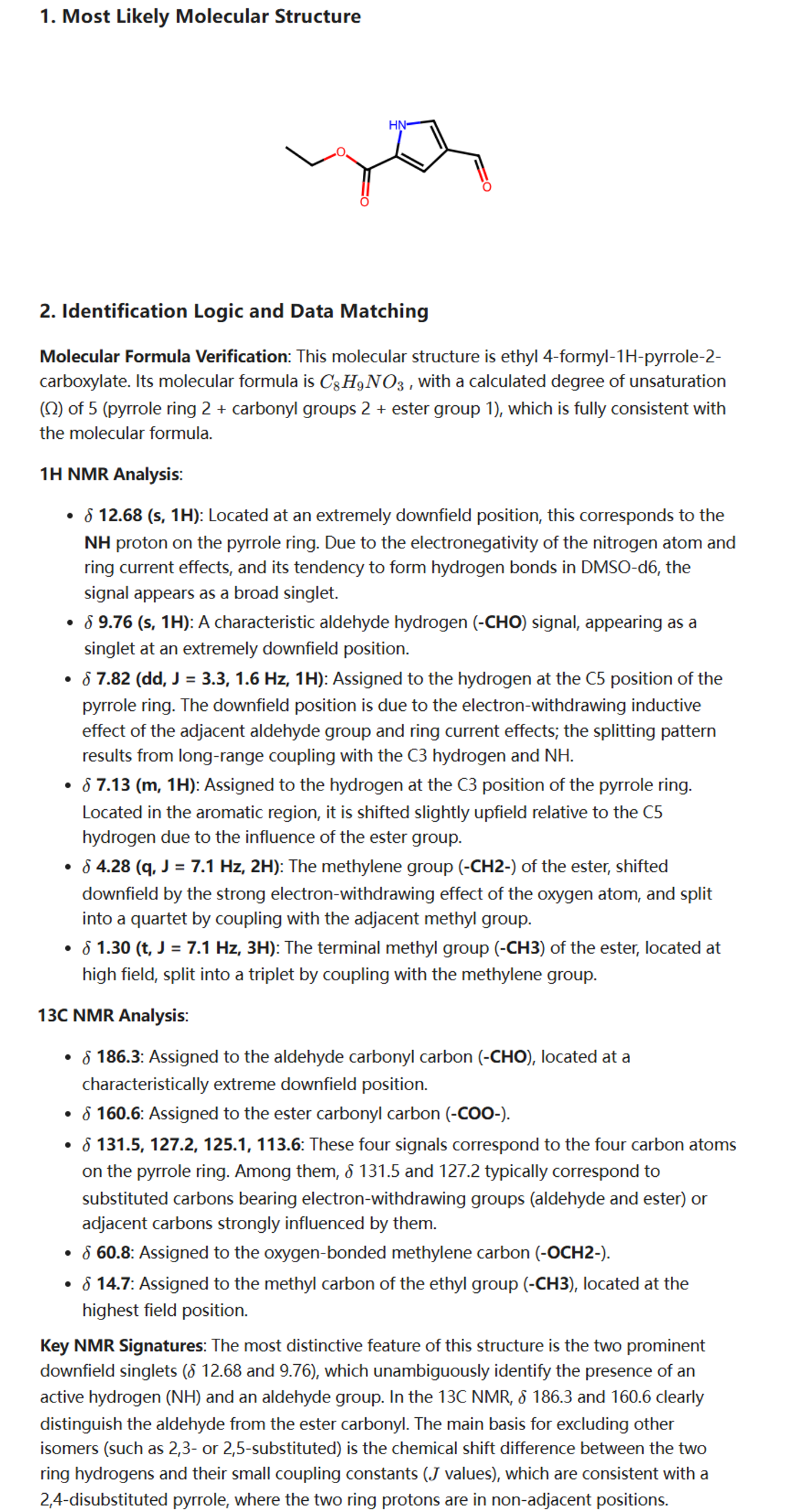}
\end{center}

\clearpage
\section{Prompt Template for the Open 500-Case Evaluation}
\label{app:open500_prompt}

For the direct-prompt baselines in Section~\ref{sec:open500_comparison}, we use the following prompt template. The fields in angle brackets are instantiated with the molecular formula, $^{13}$C NMR peaks, and $^{1}$H NMR peaks of each evaluation example.

\begin{verbatim}
You are an expert organic chemist specializing in molecular structure
elucidation from spectroscopic data. Your task is to determine the most
likely molecular structure from the provided spectroscopic information.

Instructions:
1. Do not infer the structure from known literature examples or common
   synthetic intermediates
2. Base your conclusion solely on the spectroscopic evidence provided
3. Analyze the data systematically
4. Build the structure logically

Required Output Format:

<<<FINAL_ANSWER>>>
SMILES string
<<<END>>>

The spectral information is as follows:
Molecular Formula: <MOLECULAR_FORMULA>
13C NMR Peaks: <13C_NMR_PEAKS>
1H NMR Peaks: <1H_NMR_PEAKS>
\end{verbatim}

\end{document}